\documentclass[times,twocolumn,final]{elsarticle}

\usepackage{prletters}
\usepackage{framed,multirow}
\usepackage{hyperref}

\usepackage{amssymb}
\usepackage{amsmath} 
\usepackage{latexsym}
\usepackage{fancyhdr}
\usepackage{url}
\usepackage{xcolor}
\definecolor{newcolor}{rgb}{.8,.349,.1}
\usepackage{lineno}
\usepackage{subcaption}
\journal{Pattern Recognition Letters}

\begin{document}

\ifpreprint
  \setcounter{page}{1}
\else
  \setcounter{page}{1}
\fi

\begin{frontmatter}
% \title{TREY: A joint network for handwritten text localization transcription and named entity recognition in scanned full pages.}
\title{A Neural Model for Text Localization, Transcription and Named Entity Recognition in Full Pages} % SIN PUNTO FINAL
%\title{Localize, Tag, Transcribe: A network for information extraction from scanned full pages}

%Localize, transcribe, annotate: End-to-end information extraction from full pages

\author[1,2]{Manuel Carbonell} 
\cortext[cor1]{Corresponding author: 
  Tel.: +34-93-581-3037;  
  fax: +34-93-581-1670;}
\ead{mcarbonell@cvc.uab.es}
\author[1]{Alicia Forn\'{e}s}
\author[2]{Mauricio Villegas}
\author[1]{Josep Llad\'{o}s}
\address[1]{Computer Vision Center, Computer Science Department, Universitat Aut\`onoma de Barcelona, Spain}
\address[2]{omni:us, Berlin, Germany}

\received{1 May 2013}
\finalform{10 May 2013}
\accepted{13 May 2013}
\availableonline{15 May 2013}
\communicated{S. Sarkar}

\begin{abstract}

In the last years, the consolidation of deep neural network architectures for information extraction in document images has brought big improvements in the performance of each of the tasks involved in this process, consisting of text localization, transcription, and named entity recognition. However, this process is traditionally performed with separate methods for each task.  In this work we propose an end-to-end model that combines a one stage object detection network with branches for the recognition of text and named entities respectively in a way that shared features can be learned simultaneously from the training error of each of the tasks. By doing so the model jointly performs handwritten text detection, transcription, and named entity recognition at page level with a single feed forward step. We exhaustively evaluate our approach on different datasets, discussing its advantages and limitations compared to sequential approaches. The results show that the model is capable of benefiting from shared features for simultaneously solving interdependent tasks.
\end{abstract}

\begin{keyword}
\MSC 41A05\sep 41A10\sep 65D05\sep 65D17
\KWD document image analysis \sep Information extraction \sep Text detection \sep Handwritten text recognition \sep named entity recognition \sep deep neural networks \sep multi-task learning
%% MSC codes here, in the form: \MSC code \sep code
%% or \MSC[200s8] code \sep code (2000 is the default)
\end{keyword}

\end{frontmatter}

\section{Introduction} \label{s:intro}
% \textcolor{magenta}{Add intro on document analysis/handwriting recognition (this is not an ICDAR-related journal). For example (rewrite this): There are many document collections that are only scanned, not transcribed/indexed... a manual processing is unfeasible... so document image analysis and recognition techniques applied to information extraction are needed}.
The process of information extraction from document images consists in transcribing textual contents and classifying them into semantic categories (i.e. named entities). It is a necessary process in digital mailroom applications in business documents, or record linkage in historical manuscripts. Information extraction involves localizing, transcribing and annotating text, and varies from one domain to another. Despite the recent improvements in neural network architectures, efficient information extraction from unstructured or semi-structured document images remains a challenge and human intervention is still required \cite{Toledo2019InformationEF},\cite{PalmICDAR2019}.

% \textcolor{magenta}{Here, we should mention that our WML-ICDAR paper already explored joining detection+transcription. And here, in this paper, we go further and explore the 3 tasks together.}

%  This dilemma is defined by the Sayre's paradox: good recognition of text requires a correct segmentation and vice-versa.  

In the particular case of handwritten text interpretation, the performance of handwritten text recognition (HTR) is strongly conditioned by the accuracy of a previous segmentation step. But in the other way around, a good segmentation performance can be boosted if the words are recognized. This chicken-and-egg problem (namely \emph{Sayre's paradox}) can also be stated for other steps in the information extraction pipeline: transcription vs named entity recognition (NER), or localization vs named entity categorization when there is an inherent positional structure in the document (e.g. census records, invoices or registration forms).

%This fact motivates us to formulate the hypothesis of our work. It makes sense to explore joint models that combine some of these tasks and perform information extraction in an end-to-end architecture. 
This fact motivates us to hypothesize that joint models may be beneficial for information extraction. In \cite{carbonell_jointner} we studied the combination of handwritten text recognition with named entity recognition in text lines. Later on, we explored in \cite{carbonell_wml} the interaction of text localization and recognition in full pages. In both works, we observed a benefit when leveraging the dependency of these pairs of tasks with a single joint model. 
 
%  For this reason, it makes sense to explore the combination of the three tasks in information extraction all together.

In this work we go a step further and explore the combination of the three tasks for information extraction in full pages by unifying the whole process in a single end-to-end architecture. We test our method on different scenarios, including data sets in which there is bi-dimensional contextual relevant information for the named entity tag, or there is an inherent syntactic structure in the document. Thus, we explore the benefits and limitations of an end-to-end model in comparison with architectures that integrate the different tasks of the pipeline as stand alone components. We experimentally validate the different alternatives considering different kind of documents, in particular how relevant is the geometric context, how regular is the layout, how is the strength of the named entities in the document, etc.

As far as we know, this is the first method that performs end-to-end information extraction in full handwritten pages. Our joint model is able to exploit benefits from task interaction in cases where these are strongly interdependent. Another strength of the method is its versatility, as it can be used in a wide variety of information extraction scenarios. Finally, we also contribute with a baseline for full-page information extraction in semi-structured heterogeneous documents of the IEHHR competition dataset \cite{competition}.

The paper is organized as follows. In section \ref{s:related_work} we overview the related work. In section \ref{s:method} we describe our joint model. In sections \ref{s:datasets} and \ref{s:experiments}, we present the datasets, the experimental results and discuss the advantages and limitations of our joint model. Finally, in section \ref{s:conclusion} we draw the conclusions.%, giving some recommendations for the application of joint architectures.

\section{Related Work} \label{s:related_work}
Since information extraction implies localizing, transcribing and recognizing named entities in text, we review works that deal with each of these parts separately and with pairwise task unified models.

Text localization, which can be faced as an object detection problem, has been divided into two main type of paradigms, one-stage and two-stage. In \cite{Uijlings13} a two stage method is proposed by first generating a sparse set of candidate proposals followed by a second stage that classifies the proposals into different classes and background. Regions with CNN features (R-CNN) \cite{DBLP:journals/corr/Girshick15} replaced the second stage with a CNN, improving the previous methods. The next big improvement in terms of performance and speed came with Faster-RCNN \cite{DBLP:journals/corr/Girshick15}, where the concept of \textit{anchors} was introduced.
When prioritizing speed in front of accuracy, we find one stage detection as the best option. Concretely, SSD \cite{DBLP:journals/corr/LiuAESR15} and YOLO \cite{DBLP:journals/corr/RedmonDGF15} have put one-stage methods close to two-stage in precision but having much greater speed performance. The decrease in precision of one-stage against two-stage methods is due to the class imbalance in \cite{focal_loss}, so focal loss is introduced to cope with this problem and achieve state of the art performance both in accuracy and speed. 

% \textcolor{magenta}{(Repe, reducir:)}
Regarding the transcription part, many HTR methods already perform a joint segmentation and recognition at line level to cope with the \emph{Sayre's paradox}. In this way, they can avoid the segmentation at character or word level. However, this is only partially solving the segmentation problem, because lines that are not properly segmented obviously affect the recognition. For this reason, some recent approaches propose to recognize text at paragraph level \cite{Bluche2017ScanAA}, \cite{puigcerver2017multidimensional}. But still, an inaccurate segmentation into paragraphs will affect the HTR performance. 

Taking into account those considerations, a joint method that can perform both tasks allows the noise in the predicted segmentation and obtains better transcriptions. In our previous work \cite{carbonell_wml} we proposed a model that predicts text boxes together with their transcription directly from the full page, by applying RoI pooling to shared convolutional features. In this way we avoid the need of having a perfect segmentation to get a good transcription at word level. It must be mentioned that in \cite{Wigington_2018_ECCV} the benefits of end-to-end learning for full page text recognition are put in doubt, since the best transcription performance is achieved by detecting the start of the text line, segmenting it with a line follower and then transcribing it with three separately trained networks. Nevertheless no results are shown regarding the multi-task end-to-end trained model to draw a definitive conclusion. 
% \textcolor{magenta}{FALTA comentar el ECCV 'start-follow-read'}

There are several approaches for named entity recognition. In the scenario where we have error-free raw digitized text and the goal is to sequentially label entities, most approaches \cite{Lample2016NeuralAF}, \cite{ma-hovy-2016-end}, \cite{akbik-etal-2018-contextual} use stacked long short-term memory network layers (LSTM) to recognize sequential word patterns and a conditional random field (CRF) to predict tags for each time step hidden state. Also character level word representations capture morphological and orthographic information that combined with the sequential word information achieve good results.

In the previously mentioned cases, error-free raw text is assumed for named entity recognition. In case text is extracted from scanned documents, the situation changes. In \cite{spp_toledo} a single convolutional neural network (CNN) is used to directly classify word images into different categories skipping the recognition step. This approach, however, does not make use of the context surrounding the word to be classified, which might be critical to correctly predict named entity tags.
In \cite{Rowtula18} and \cite{Toledo2019InformationEF} a CNN is combined with a \textit{Long short-term memory} (LSTM) network to classify entities sequentially thereby making use of the context, achieving good results. This is improved in \cite{carbonell_jointner} and \cite{wigington2019} by joining the tasks of text recognition and named entity recognition by minimizing the Connectionist Temporal Classification loss (CTC) \cite{Graves:2006:CTC:1143844.1143891} for both. Still, in these works there is no bi-dimensional context pattern analysis.
Very recently an attention-based method \cite{guo2019eaten} performs entity extraction in a very controlled scenario as ID cards, where a static layout implies that is not necessary to detect complex text bounding boxes. In summary, all these works suggest that it is promising to explore methodologies that integrate the three tasks in a unified model.

% \textcolor{magenta}{Falta poner alguna frase de resumen del estado del arte que enlace con nuestro metodo.}

% \textcolor{magenta}{Manu, se tiene que mencionar en la intro que nos beneficiamos del 2D context: }\textcolor{blue}{

% The limitation in those cases is that in case of having bi-dimensional spatial relationships these cannot be recognized by the model, which motivates our proposed method.
% }

% \textcolor{magenta}{.
% Icdar cwigington multi ctc
% }

% \textcolor{blue}{POSA https://arxiv.org/pdf/1909.09380.pdf}
\section{Methodology} \label{s:method} 
\begin{figure*}
    \centering
    \includegraphics[width=\textwidth]{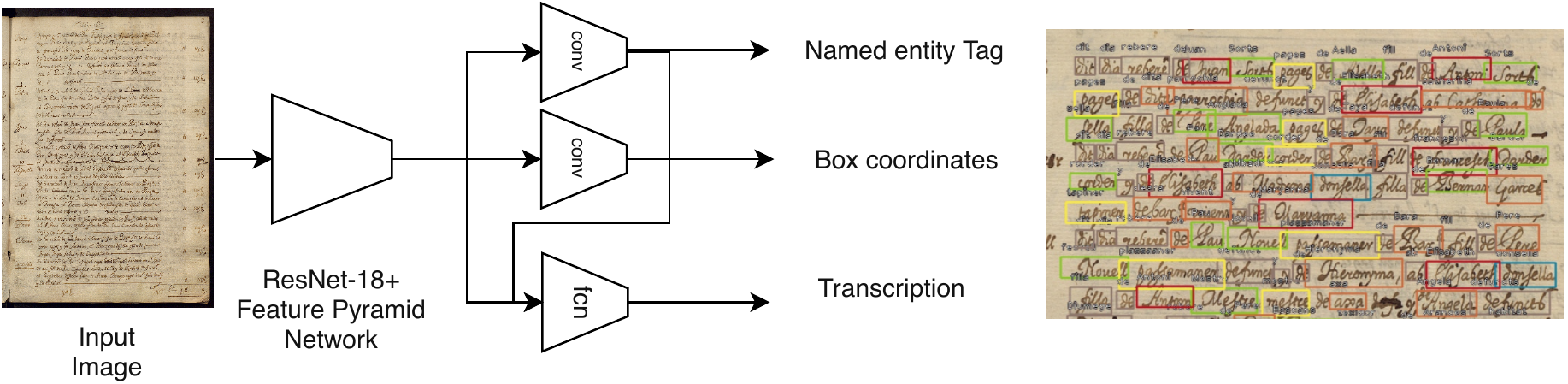}
  
    \caption{Overview of the proposed method. Convolutional features are extracted with ResNet18 and FPN. The classification and regression branches calculate the positive boxes and the recognition branch predicts the transcription of the content of each box. Cross entropy, squared-sum and CTC losses are backpropagated through the whole model for training.}
    \label{fig:overview}
\end{figure*}

As introduced before, our model extracts information in a unified way. First, convolutional features are extracted from the page image, and then, different branches analyze these features for the tasks of classification, localization, and named entity recognition, respectively. An overview of the architecture is shown in Figure \ref{fig:overview}. 

\subsection{Shared features}
Since the extracted features must be used for very different tasks, i.e. localization, transcription and named entity recognition, we need a deeper architecture than the one used for each isolated task. According to recent work on object detection and text semantic segmentation \cite{DBLP:journals/corr/LinDGHHB16}, a proper architecture to extract convolutional features from the image is Residual Network 18 (ResNet18) \cite{DBLP:journals/corr/HeZRS15}. We have considered exploring deeper architectures or variations of ResNet18 to improve the final performance, yet the scope of this work is not to find the best feature extraction but to unify the whole information extraction process.
ResNet18 consists of 5 residual convolutional blocks. Each of those encloses 2 convolutional layers, followed by a rectifier linear unit activation and a residual connection. Table \ref{resnet18} shows the detailed list of blocks and configuration of the shared feature extractor.
% An illustration of this can be seen in Figure \ref{resnet_block}. 
%The detailed list of blocks and their configuration is shown in Table \ref{resnet18}.

% \subsection{Feature extractor}
% The first module of our model is a deep feature extractor, whose weights are shared for the recognition and detection tasks. Taking into account that the localization of text in a scanned document (where we are previously aware of its existence) might be easier than detecting an object in the wild, we have chosen the ResNet18 \cite{DBLP:journals/corr/HeZRS15}, a light state-of-the-art architecture for object detection and classification.  

% \begin{figure}[h]
% \centering
% \includegraphics[width=0.2\textwidth]{resnet_block.png}
% \caption{A ResNet18 convolutional block of 64 channels. Equivalent blocks are added with 128, 256 and 512 channels.}
% \label{resnet_block}
% \end{figure}

\begin{table}[h]
\center
\caption{ResNet18 architecture used for feature extraction.}
\label{resnet18}
\begin{tabular}{llll}
\hline
\textbf{Layer}   & \textbf{output shape} & \textbf{kernel size} & \textbf{$\mid$ kers $\mid$} \\ \hline
res-conv-block 1 & H/2$\cdot$W/2         & 3 x 3                & 64                          \\
res-conv-block 2 & H/4$\cdot$W/4         & 3 x 3                & 64                          \\
res-conv-block 3 & H/8$\cdot$W/8         & 3 x 3                & 128                         \\
res-conv-block 4 & H/16$\cdot$W/16       & 3 x 3                & 256                         \\
res-conv-block 5 & H/32$\cdot$W/32       & 3 x 3                & 512                        
\end{tabular}
\end{table}

% \begin{figure}
%     \centering
%     \includegraphics[width=0.4\textwidth]{det_ner_seq.png}
%     \caption{} %Architecture details can be found in table \ref{table:architecture}}
%     \label{fig:fpn}
% \end{figure}

\begin{figure}[tbh]
    \centering

    \setlength\fboxsep{0pt}
    \begin{tabular}{c}
    \includegraphics[width=0.9\linewidth]{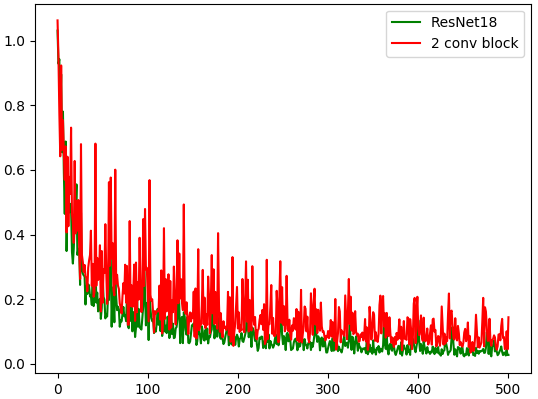} \\[2mm]
    \fbox{\includegraphics[width=\linewidth]{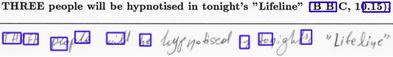}} \\[2mm]
    \fbox{\includegraphics[width=\linewidth]{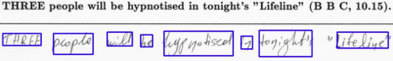}} \\
    \end{tabular}

    \caption{(Top) Plot comparing the regression loss during training on the benchmark dataset IAM for ResNet18 (green) against a 2 convolutional block reduced version (red). (Middle) Predictions on IAM with 2 convolutional block model. (Bottom) Predictions on IAM with ResNet18 model. The model does not confuse relevant with irrelevant text.}
    \label{fig:light_heavy}
\end{figure}

%\textcolor{olive}{ResNet-18 o ResNet18? usa la misma forma en todo el paper}, 
We have also tried to reduce the amount of layers of the ResNet18 but when doing so, the training on the benchmark text recognition dataset IAM \cite{Marti2002TheIA} converged slower and the final validation error was higher than when using the full ResNet18 architecture. We attribute this effect to the increase of noisy detections and false positives, in which the model was confusing relevant with non-relevant text. A visualization of the ResNet18 training compared to a 2 block lighter architecture and the corresponding predictions after 20 epochs can be seen in Figure \ref{fig:light_heavy}.

Consequently, we chose an intermediate depth model which allows to tackle such complex tasks at once. The output of the Feature Pyramid Network is a set of 5 down sampled feature maps with scales 1/8,1/16,1/32,1/64,1/128 with respect to the input image. Each of these are forwarded to the upcoming branches and their output is stacked in a single tensor, from which we later select the most confident predictions.
% We have tried other even lighter configurations, but when reducing the amount of layers, we observed slower convergence and worse final results. This was most probably caused by noisy detections and false positives, confusing text with non-relevant text. For this reason we have chosen an intermediate depth architecture that allowed regressing the characteristics of the text, and skipping the step of separating the regions of interest (i.e. including the layout analysis step in the whole process). 

% %For this reason we have chosen to not build a even lighter architecture, as ResNet18 allowed to successfully regress the regions of interest 

% Based on recent work on object detection, we build a feature pyramid network (FPN) \cite{DBLP:journals/corr/LinDGHHB16}  which combines the extracted deep features of different levels of abstraction by means of \textit{deconvolutions}. These type of layers consist of bilinear interpolation to apply differentiable upscaling of the high level features, followed by convolutions to reduce the number of channels, allowing to add them to lower level features. A diagram of this module is shown in Figure \ref{fig:fpn}. This approach gave a boost in performance to detect objects at different scales, which is definitely also a beneficial feature for localizing text in documents.

\subsection{Classification branch}
\label{class_branch}
Concerning the detection and classification of objects in images, there are different approaches regarding the prediction of the probability of an object being present in a given location of the image. The two most used options are, either to regress the intersection over union (IoU) of the predicted box with the ground truth box as done in \cite{DBLP:journals/corr/RedmonDGF15}, or to predict the probability of each object of a given class for each location with a separate branch, encoded as a one-hot vector as in \cite{DBLP:journals/corr/LinDGHHB16}. We have chosen this second option due to its performance for a wide variety of data sets. The architecture of this branch is shown in table \ref{classbranch}.

\begin{table}[h]
\center
\caption{Classification and regression branch architectures, where downsampling levels are $ds_{l_i}\in\{8,16,32,64,128\}$.}
\label{classbranch}
\begin{tabular}{llll}
\hline
\textbf{Layer} & \textbf{output shape}                                  & \textbf{kernel size} & \textbf{$\mid$ kers$\mid$} \\ \hline
conv-block 1   & $ H/ds_{l_i}\cdot W/ ds_{l_i} $                        & 3 x 3                & 256                \\
conv-block 2   & $H/  ds_{l_i}\cdot W/ds_{l_i}  $                       & 3 x 3                & 256                \\
conv-block 3   & $ H/ds_{l_i}\cdot W/ds_{l_i}  $                        & 3 x 3                & 256                \\
conv-block 4   & $H/ds_{l_i}\cdot W/ds_{l_i}$                           & 3 x 3                & 256                \\
conv-block 5   & $n_{\text{anchors}} \cdot \{n_{\text{classes}} ,4\}  $ & 3 x 3                & 1                 
\end{tabular}
%\caption{Architecture of the objectness and regression branches applied to downsampled convolutional features for each level $l_i$ of the feature pyramid.}
%\textcolor{olive}{Tabla con 2 captions? Y este segundo caption es mas como texto que un caption.}
\end{table}

We also explored to use this branch as a named entity classifier. The motivation behind is to take context into account through the prediction of the presence of certain features in a neighbourhood of a point of the convolutional grid. The difficult part comes when attempting to capture dependencies between distant parts of the image, as it happens when a sequential approach is used. More specifically, the classification branch, or objectness loss in case of a pure text localizer classifier, is trained with the following cross-entropy loss function:

\begin{equation}
CE(p_{cl}) = -(y_{cl} \cdot log(p_{cl}) + (1 - y_{cl}) \cdot log(1 - p_{cl}))
\label{eq:cross_entropy}
\end{equation}

% \subsection{Classification branch}
% For each one of the levels of the pyramid, the extracted features are fed to the classification network, which after four convolutions with rectifier linear unit and a final sigmoid layer, will predict the probability of the presence of a word and its category, i.e. we tackle the word localization and named entity recognition task as object detection and classification.

% After predicting the class probabilities vector $p_{cl}$, the binary cross entropy (CE) loss is calculated and backpropagated through the classification branch and shared feature network. Formally, CE is computed as follows:

\subsection{Regression branch}
\begin{figure}
\centering
\includegraphics[scale=0.45]{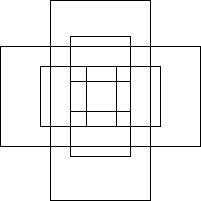}
\caption{Anchors combining ratios $\frac{1}{2},1,2$ and scales $1$, $2^\frac{1}{3}$, $2^\frac{2}{3}$, with a base size of $32$ (9 anchors).}% as shown in figure \ref{anchors}}
\label{anchors2}
\end{figure}

To predict the coordinates of the box positives, the regression branch receives the shared features and, after 4 convolutions with rectifier linear unit (ReLU), it predicts the offset values of the predefined anchors. Formally: 
% In a similar way, the regression network receives the whole page image features and after four convolutions, it regresses the box coordinate offsets from the predefined anchors. Formally: 

\begin{align}
\centering
x &= X +{ dx} \cdot W\nonumber \\
y&= Y + {dy} \cdot H\nonumber\\
w&=e^{dw} \cdot W\nonumber\\
h&= e^{dh }\cdot H \label{box_regress}
\end{align}
where $(x,y,w,h)$ are the predicted box coordinates, $dx, dy, dw, dh$ are the predicted offsets and $X, Y, W, H$ are the predefined anchor box values. 
The offset of the predefined anchors is regressed by minimizing the mean square error:
\begin{equation}
    MSE(\delta ,\delta ')=(\delta - \delta')^2
\end{equation}
being $\delta$ the vector of target offsets of the anchors for each ground truth box.
The anchors are generated as the combination of the ratios $\frac{1}{2}$,1,2 and the scales 1, $2^\frac{1}{3}$, $2^\frac{2}{3}$ with a base size of 32 (9 anchors) as shown in Figure \ref{anchors2}.

\subsection{Feature pooling}
Once we have predicted the class probabilities and the coordinate offsets for each anchor in each point of the $Im_H/8\times Im_W/8$ convolutional grid, we select the boxes whose confidence score surpasses a given threshold, and remove the overlapping ones applying a non-maximal suppression algorithm. With the given box coordinates, we apply RoI pooling \cite{DBLP:journals/corr/Girshick15} to the convolutional features of the full page, but saving the input to allow error backpropagation from further branches. We use the 5 levels of the feature pyramid to calculate the box anchor offsets and the objectness values. Conversely, for computational reasons, we only keep the least downsampled features for the text recognition and named entity recognition branches, as we need the highest possible resolution for those tasks. 

\subsection{Text recognition branch}
As in \cite{carbonell_wml} we build a recognition branch that will predict the text contained in each box. The architecture of this branch, shown in Table \ref{recogbranch}, consists of two convolutional blocks followed by a fully connected layer. The output of this layer is the probability of a character for each column of each one of the pooled features. From these predictions, we calculate the Connectionist Temporal Classification (CTC) loss \cite{Graves:2006:CTC:1143844.1143891}. This loss, or Maximum Likelihood error, is calculated as the negative logarithm of the probability of a ground truth text sequence given the network outputs 

\begin{equation}
O^{ML}(\mathcal{D},Y')=-\sum_{(x,y)\in D}ln(p(y|x))   
\end{equation}being $\mathcal{D}={(x,y)_{i=1}^{i=N}}$ the set of training input-target pairs and $Y'$ the set of network sequence outputs.

To calculate this probability, we add the probabilities of all possible paths in the [time steps $\times$ alphabet length] matrix with the forward-backward algorithm. Repeated output characters not separated by blank (non-character) are joined using the collapse function $\mathcal{B}$, for example, $\mathcal{B}$(hheeel-llo)=hello.
This loss is added to the classification and regression losses to backpropagate them together for each gradient update.

\begin{table}[h]
\center
\caption{Recognition branch architecture.}
\label{recogbranch}
\begin{tabular}{llll}
\hline
\textbf{Layer}  & \textbf{Output shape}        & \textbf{ker size} & \textbf{$\mid$ kers$\mid$} \\ \hline
conv-block 1    & pool H$\cdot$pool W          & 3 x 3             & 256              \\
conv-block 2    & pool H$\cdot$pool W          & 3 x 3             & 256              \\
%BLSTM-1         & pool H$\cdot$pool W          & 256               & -                \\
%BLSTM-2         & pool H$\cdot$pool W          & 256               & -                \\
Fully connected & pool W$\cdot$  $|alphabet| $ & -                 & -               
\end{tabular}
\end{table}

\subsection{Semantic annotation branch}
\label{semantic_branch}
As we mentioned before, one possibility to assign a semantic tag to each word is to predict its class from the classification branch for each anchor. However, this would not capture the context as the activations only rely on the convolutional feature maps of a neighborhood of each point. For this reason, we add this network branch to predict the semantic tags as a sequence from the ordered pooled features of each box. For simple layouts, such as single paragraph pages, the pooled features, which correspond to text boxes in the page, are sorted in reading order (i.e. left to right and top to bottom) by projecting a continuation of the right side of the text box.  
Once we have the ordered pooled features, we pad them and apply two convolutions followed by a fully connected network as a standard named entity recognition architecture. Then, we minimize the cross entropy loss shown in equation \ref{eq:cross_entropy} for each of the sequence values.

%\textcolor{olive}{left-to-right y top-to-bottom es el reading order pero solo para layouts simples, como un único párrafo, debes dejar esto claro o evitar decir reading order}.

% \subsection{Box Sampler}
% Ideally we would like the end-to-end model to base the positives and negatives of box candidates to take in account the text recognition score. Nevertheless, doing this calculation for each anchor would be prohibitively expensive, as we 

\subsection{Receptive field calculation}
Our approach assumes that each activation of a neuron in the deepest layers of a CNN depends on the values of a wide region of the input image, i.e. its receptive field. Also it is important to notice that the closer a pixel is to the center of the field, the more it contributes to the calculation of the output activation. This can be a useful property for documents where the neighboring words determine the tag of a given word, but it can also be a limitation when distant entities are related in a document. To calculate how much context is taken in account for each unit of the features that are fed to the RoI pooling layer, we must look at the convolutional kernel sizes $k$ and strides $s$ of each layer. In this way, as in \cite{Dumoulin2016AGT}, we can calculate the relation between the receptive field size of a feature map depending on the previous layer's feature map:

\begin{equation}
    r_{out}=r_{in}+(k-1)\cdot j_{in}
\end{equation}
where $j_in$ is the \textit{jump} in the output feature map, which increases in every layer by a factor of the \textit{stride}
\begin{equation}
    j_{out}=j_{in}*s
\end{equation}

By using these expressions with our architecture (ResNet 18 + FPN), we obtain a receptive field size of 1559 in the shared convolutional feature map. That means that, since the input images are 1250$\times$1760, the values predicted for each unit mostly depend on the content of the whole page, giving more importance to the corresponding location of the receptive field center.

\section{Datasets} \label{s:datasets}

One of the limitations when exploring learning approaches for information extraction is the few publicly available annotated datasets, probably due to the confidential nature of this kind of data. Nonetheless, we test our approach on three data sets. The details of amount of pages, words, out of vocabulary (OOV) words and partitions can be found in Table \ref{dataset_details}.

%\textcolor{magenta}{Manu, pon info de num.writers, printed/hw/mixed, structured/no-structured... en la tabla 5} Lo de writer info lo pongo en la descripcion del dataset escrita ya que no hay información específica para todos los datasets

\begin{table}[h]
\centering
\caption{Characteristics of the datasets used in our experiments. Entities refer to the amount of relevant words (i.e. they do not belong to the class "other").}
\label{dataset_details}
\begin{tabular}{lllll}
\hline
\textbf{}              & \textbf{Part} & \textbf{IEHHR} & \textbf{WR} & \textbf{sGMB} \\ \hline
\multirow{3}{*}{Pages} & train         & 79             & 994         & 490           \\
                       & valid         & 21             & 231         & 53            \\
                       & test          & 25             & 323         & 50            \\ \hline
\multirow{3}{*}{Words} & train         & 2100           & 2837        & 7010          \\
                       & valid         & 878            & 731         & 1740          \\
                       & test          & 1020           & 1033        & 4085          \\ \hline
$\|$OOV$\|$            & all           & 387            & 853         & 1372          \\
\% OOV                 & all           & 37             & 82          & 34 \\ \hline
      
      \% entities & all & 52.5 & 100& 17\\ \hline 
      $\|$entity tags$\|$& - & 5 &3 & 5 \\ \hline 
\end{tabular}
\end{table}

\subsection{IEHHR}
\begin{figure}[tb]
    \centering
    \includegraphics[width=0.45\textwidth]{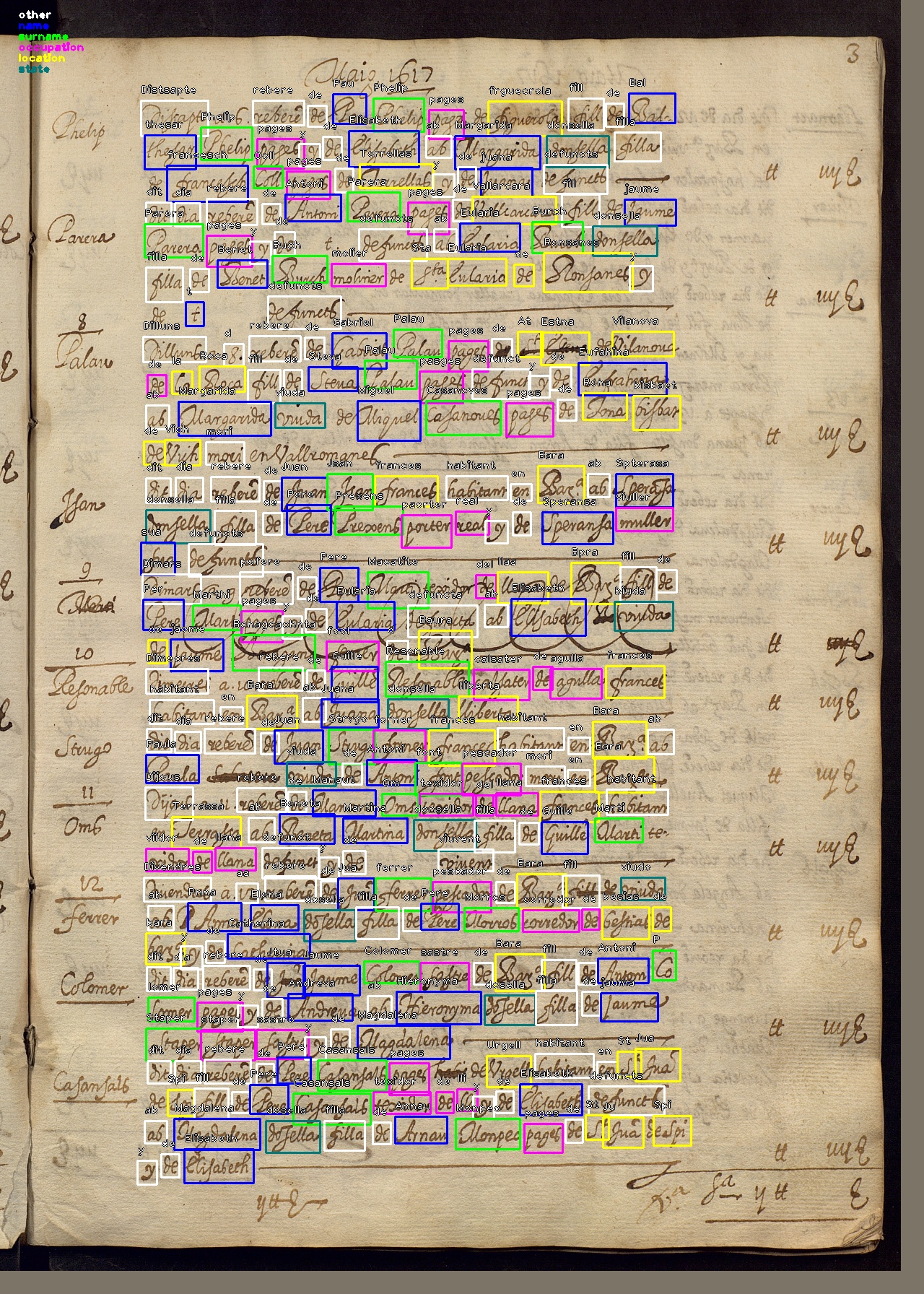}
    \caption{Word localizations, transcriptions and semantic annotations on an unseen page of the IEHHR dataset. The model learns to detect and classify words based not only on its appearance but also on its context. The colors illustrate the different type of named entities.}
    \label{fig:full_page}
\end{figure}

The IEHHR dataset is a subset of the Esposalles dataset \cite{competition} that has been labeled for information extraction, and contains 125 handwritten pages with 1221 marriage records (paragraphs). Each record is composed of several text lines giving information of the husband, wife and their parents' names, occupations, locations and civil states. On the sides of each paragraph we find the husband's family name and the fees paid for the marriage. An example page is shown in Figure \ref{fig:full_page}. 

\subsection{War Refugees}
\begin{figure}[t]
    \centering
    \includegraphics[width=0.20\textwidth]{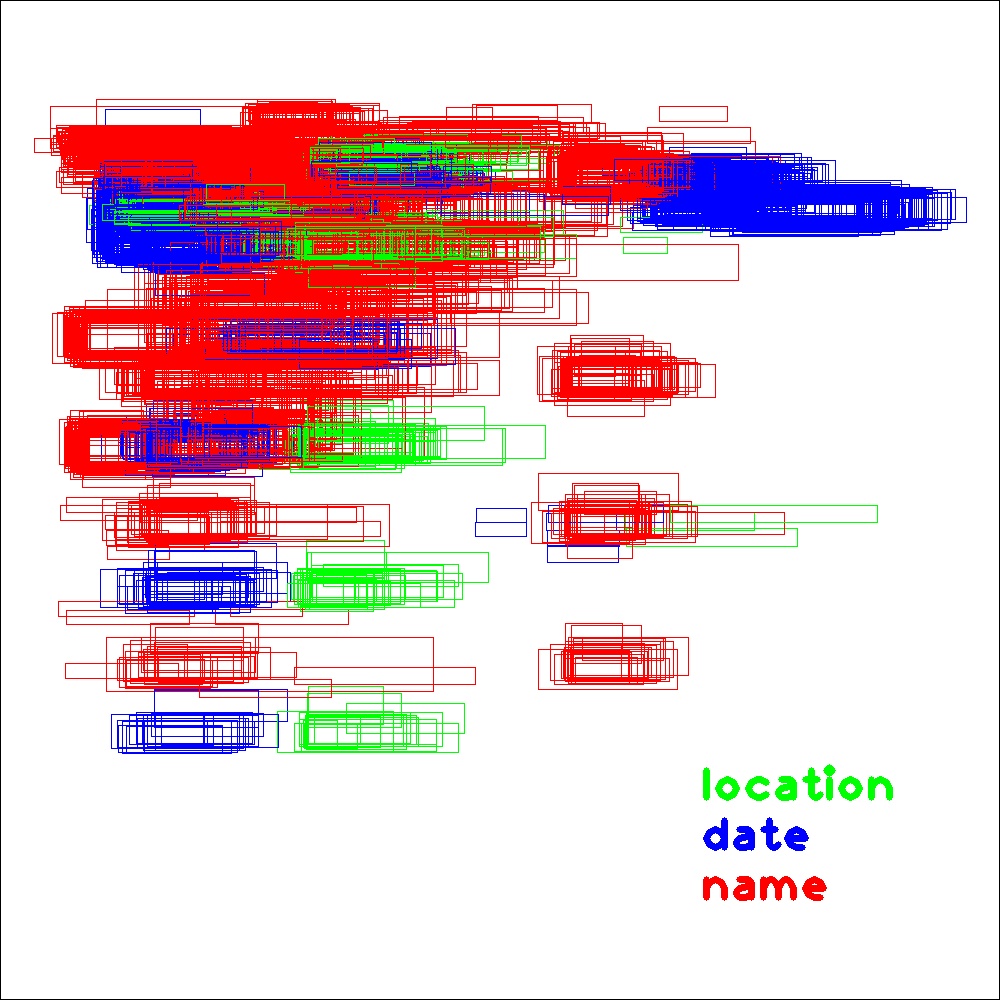}
    \caption{Normalized bounding boxes of the tagged text of all training images in the WR dataset.}
    \label{fig:arolsen}
\end{figure}

%\textcolor{olive}{Mejor que sea War Refugees (WR)}
The War Refugees (WR) archives contain registration forms from refugee camps, concentration camps, hospitals and other institutions, from the first half of 20th century. We have manually annotated the bounding boxes, transcriptions and entity tags of names, locations and dates. Due to data privacy we cannot share the images, but instead we show in Figure \ref{fig:arolsen} a plot of all annotated text normalized bounding boxes, where the colors correspond to different tags. As we can observe, there is a strong pattern relating the text location and its tag, although it is not fixed enough for applying a template alignment method. The main difficulty of this dataset is to distinguish relevant from non-relevant text, which in most cases only differs by its location or by a nearby printed text key description. Another challenge is the high amount (82$\%$) of out of vocabulary words, together with the high variability of the writing style and the mixture of printed and handwritten text. 

\subsection{Synthetic GMB}
\begin{figure}[t]
    \centering
    \includegraphics[width=0.5\textwidth]{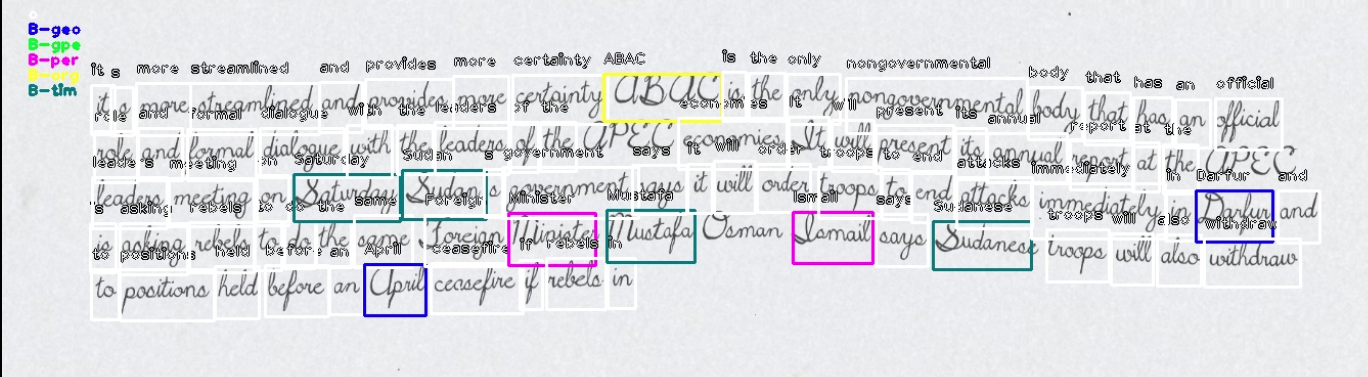}
    \caption{A generated page from the SynthGMB dataset. A major difficulty is the sparsity of named entities with respect the other words.}
    \label{fig:sgmb}
\end{figure}

We have generated a synthetic dataset (sGMB) to explore the limitations of our model, concretely in a standard named entity recognition task, in which text is unstructured and the amount of named entities within the text is low. For this purpose, we have generated synthetic handwritten pages with the text of the GMB dataset \cite{Bos2017GMB} by using synthetic handwritten fonts, applying random distortions and noise to emulate realistic scanned documents. Although it is easier to recognize synthetic documents than real ones, the difficulty here remains on the sequential named entity recognition task, especially because, contrary to the previous datasets, here the text does not follow any structure. An example is shown in Figure \ref{fig:sgmb}. The dataset and ground truth are available here
\footnote{\url{https://github.com/omni-us/research-dataset-sGMB}}.

{}

% \subsection{FUNSD}
% This dataset consists of 199 noisy scanned forms of containing a subset of the e RVL-CDIP tobacco documents with text location, transcription and class annotations. The documents have a huge variety in appearance, making form understanding very challenging, specially taking in account the limited amount of examples for this case. As it is stated in the report, it cannot be ensured that there are enough examples to create a generic form understanding application \cite{DBLP:journals/corr/abs-1905-13538}. Still we wanted to try how our model adapts to this case.

% Please add the following required packages to your document preamble:
% \usepackage{multirow}
% Please add the following required packages to your document preamble:
% \usepackage{multirow}
%\begin{table}[]
%\caption{Characteristics of the datasets used in our experiments.}
%\label{dataset_details}
% Please add the following required packages to your document preamble:
% \usepackage{multirow}
% Please add the following required packages to your document preamble:
% \usepackage{multirow}

% \begin{table}
% \centering
% \caption{Marriage Records dataset distribution}
% \label{dataset_details}
% \begin{tabular}{llll}
%         & Train & Validation & Test \\ \hline
% Pages   & 95    & 29         & 14   \\
% Records & 872   & 253        & 96   \\
% Lines   & 2759  & 757        & 311  \\
% Words   & 28346 & 8026       & 3155 \\ \hline
% \multicolumn{4}{l}{Out of vocabulary words: 5.57 \%}      
% \end{tabular}
% \end{table}

\section{Experiments} \label{s:experiments}

In this section we describe the experiments performed for each data set. We optimized our model with stochastic gradient descent on the three described losses in section \ref{s:method}. When unifying tasks in a single model, the model becomes quite memory expensive. To overcome this limitation, during the training, we had to set our batch size to be 1 page. Based on previous object detection work we have chosen the learning rate to $10^{-4}$, the non-maximum suppression threshold to 0.2 and the box sampling score threshold to 0.5. We have chosen a patience of 100 epochs to trigger early stop.

\subsection{Setup}

In this work we propose a method for unifying the whole process of extracting information from full pages in a single model. Nevertheless, our approach has been evaluated using the following different configurations:

\begin{itemize}

\item \textbf{A: Triple task model}.
The first method variation consists in using our proposed model to perform all tasks in a unified way using the objectness branch for named entity classification as explained in section \ref{class_branch}, with no sequential layers but only convolutional ones.

\item \textbf{B: Triple task sequential model}.
The second variation also performs the three tasks in a unified way, but by concatenating the pooled features in reading order and predicting the labels sequentially, as explained in section \ref{semantic_branch}.

\item \textbf{C: Detection + named entity recognition}.
In this case we consider to face the extraction of the relevant named entities as a detection and classification problem. Here, we ignore the recognition part and only backpropagate the classification and regression losses from their respective branch outputs. We also consider the sequential version of this approach explained in section \ref{semantic_branch}.

\item \textbf{D: Detection + transcription}.
Here we combine in a unified model the tasks of localization and transcription, as seen in \cite{carbonell_wml}, in contrast to an approach in which the two tasks are faced separately, where the recognition model would cope with inaccurate text segmentations. Here we aim to observe how precisely we can obtain text boxes and transcriptions, so that named entity recognition can be applied afterwards. 

\item \textbf{CNN classifier}.
Finally, we evaluate the variability of the cropped words among the different categories and the difficulty of annotating words separately. Thus, we train a CNN network, similar to the classification branch from our proposed method, that classifies words without using any shared features for recognition or localization. So, this network does not benefit from context information.

\end{itemize}

\begin{figure}[t]
    \centering
    \includegraphics[width=0.82\linewidth]{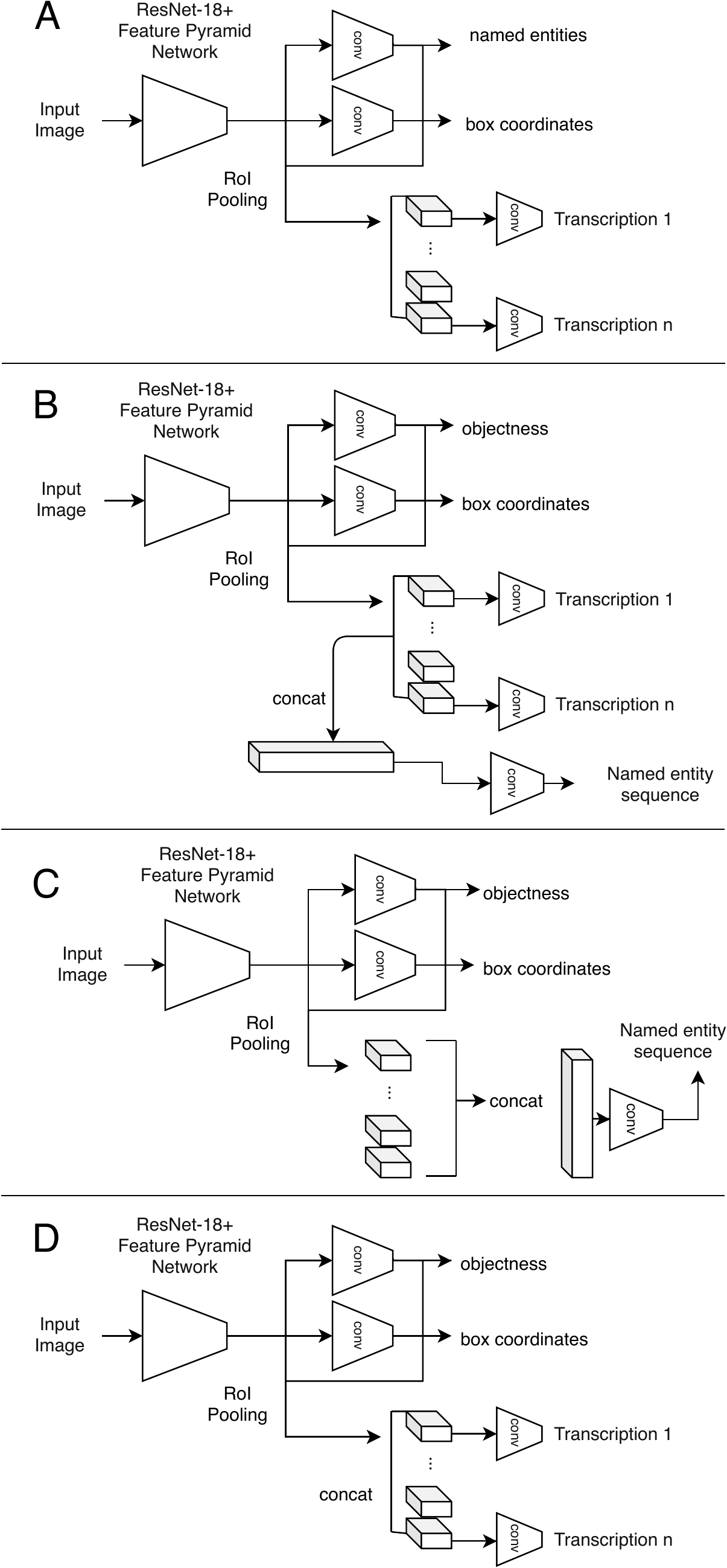}
  
    \caption{Model setup variations, A-D. The differences rely on how the named entities are recognized, and the optional integration of the recognition branch.}
    \label{fig:setups}
\end{figure}

Diagrams of each setup can be seen in Figure \ref{fig:setups}.
The full source code for all experiments is publicly available here  \footnote{\url{https://github.com/omni-us/research-e2e-pagereader}}.

\subsection{Metrics}
%\textcolor{blue}{Joan: També explicaria la mesura del MAP}

Different metrics have been used to evaluate the proposed methodology. One to evaluate the performance of the text detection, one for named entity recognition, and another for transcription. For text localization we used the the widely used object detection metric, average precision. For named entity recognition, we used the $F1$ score. Let $p=\frac{TP}{TP+FP}$ be the precision metric, i.e. the number of true positives out of the total positive detections; $r=\frac{TP}{TP+FN}$ be the recall metric, i.e. the number of true positives out of the total ground truth positives, i.e. the true positives plus false negatives.
We consider the recall-precision map, $p:[0,1]\mapsto[0,1]$ which maps the recall value $r$ to the precision $p$ that we obtain if we had the detection threshold to get such a recall. Then, the Average Precision is the value $\int_0^1 p(r)$, i.e. the area under the precision-recall graph. %\textcolor{red}{We also use this metric for evaluating the named entity recognition.}
The $F1$ score consists of the harmonic mean between the precision and the  recall:
\begin{equation}
    F1 = 2\cdot \frac{precision\cdot recall}{precision+recall}
\end{equation}
For text localization all the word bounding boxes are considered in the metric. In contrast, for named entity recognition only the entity words count, as it is standard in this kind of task. Also the recognized text is not used to match the entities, only the location of the word within the page and the entity tag. This is so that the metric only evaluates the prediction of entities completely decoupled from the transcription performance.

For the transcription score we use the Character Error Rate (CER), i.e. the number of insertions, deletions and substitutions to convert the output string into the ground-truthed one, divided by the length of the string. Formally:

\begin{equation*}
    CER =\frac{i + s + d}{\text{label length}}
\end{equation*}

\subsection{Results}

From the results shown in Table \ref{t:results}, we observe that the localization performance is high in general being this one the less challenging task of all. Differences among the different setups and datasets are not significant.

For text recognition all methods perform similarly except in WR dataset. Setup A shows a slight better performance than B and D. This suggests that the eavesdropping effect mentioned in \cite{DBLP:journals/corr/Ruder17a} might be taking place in this case. The outnumbering of non-labeled words vs labeled in WR dataset increases considerably the difficulty of the recognition task for this dataset due to the same type of error observed in figure \ref{fig:light_heavy}. We cannot show images of this phenomena observable during training due to the confidential nature of the documents. A possible solution to this would be to also have annotated words which are not entities as it happens in IEHHR and sGMB.

For the entity recognition part, we do not observe significant differences among the end-to-end approaches in the \textit{IEHHR} dataset. This suggests that the local neighborhood information seems enough to give correct predictions. The high named entity recognition performance using the triple task sequential approach (case B) in WR dataset suggests that it is beneficial to combine the tasks of named entity recognition and localization. To have a better idea of whether our method makes use of context or the sole content of the word is sufficient, we compare its performance to the CNN classifier for segmented words. Since we are facing a named entity recognition task, we only take into account the entities, i.e. words labeled as 'other' are not taken into account after the text localization step. By doing so, in the IEHHR dataset we observe a greater performance for the end-to-end models compared to the CNN. This is an evidence that context is being used since in the CNN approach there is the advantage that an explicit perfect segmentation is given. In the case of WR we observe an even greater boost of performance when using the context, specially the bi-dimensional one (setup A). We attribute this to the inherent layout pattern contained in the dataset as it can be intuited observing Figure \ref{fig:arolsen}. We also assume that predictions were based on the layout due to the large amount of out of vocabulary words, which would make it difficult to predict the word category based on a known vocabulary. sGMB is clearly the most difficult one since it has the sparsest distribution of entities as it can be seen in Table \ref{s:datasets}. With this dataset we also see a very substantial increase when using the context, concretely, the sequential one (setups B and C). This makes sense as the type of text is natural language, which means that it has a sequential structure but lacks the bi-dimensional layout structure of registration forms or marriage records. Consequently, sequential patterns existing in natural language are more suitable to recognize entities.

\begin{table}[]
\centering
\caption{Performance of the different method variations on each dataset.}
\begin{tabular}{llll}
\hline
\textbf{Method}    & \textbf{IEHHR} & \textbf{WR} & \textbf{sGMB} \\
\hline
\multicolumn{4}{c}{\textbf{Text localization (AP)}}                        \\ \hline
A: triple task     & 0.97                & 0.976            & 0.994         \\
B: triple task seq & 0.972               & 0.973            & 0.994         \\
C: Det+ner seq     & 0.969               & 0.975            &\textbf{ 0.997}         \\
D: det+htr         & \textbf{0.974 }              & \textbf{0.981  }          & 0.996  \\
\hline
\multicolumn{4}{c}{\textbf{Text recognition CER (\%)}}                      \\ \hline
A: Triple task     &\textbf{6.1}                 & \textbf{23.7}               & \textbf{2.3}          \\
B: Triple task seq & 6.3                 & 28.7               & 2.6           \\
D: Det+HTR         & 6.5                 & 27.5               & 2.5           \\       
\hline

\multicolumn{4}{c}{\textbf{Named entity recognition (F1)}}                   \\ \hline
A: Triple task     & 0.797               & \textbf{0.975}   & 0.347     \\
B: Triple task seq & \textbf{0.806}      & 0.924            & \textbf{0.535}         \\
C: Det+NER seq     & 0.796               & 0.963            & 0.510         \\ %\hline \hline
%\multicolumn{4}{c}{\textbf{Isolated Named Entity Recognition  (F1)}}             \\ \hline
CNN classifier     & 0.700               & 0.821            & 0.382          

\end{tabular}
\label{t:results}

\end{table}

\section{Discussion and Conclusion} \label{s:conclusion}
In this paper we have presented a unified neural model to extract information from semi-structured documents. Our method shows the strengths of the pairwise interaction of some of the tasks, such as localization and transcription and also for localization and named entity recognition when the spatial information or the neighbourhood (geometric context) of a text entity influences the value to predict. Nevertheless observing the performance of triple task neural model variations, it must be noted that a unified model can be limited in performance in cases where one specific task is much harder and unrelated to the others. In such a case, a separate approach would allow us to use specific techniques for this difficult unrelated task. For example, named entity recognition performance is limited by the fact that it is very difficult to generate semantically meaningful word embedding vectors (e.g. word2vec, glove) when the model input is a page image.

In summary, we conclude that a joint model is suitable for cases in which there is a strong task interdependence, but not for unstructured documents where the main difficulty is on one independent single task. 

\section*{Acknowledgments}
%\textcolor{olive}{Incluye también ``This work was supported by the European Commission H2020 SME Instrument program, under Grant Agreement no. 849628, project OMNIUS%''}

This work has been partially supported by the Spanish project RTI2018-095645-B-C21, the European Commission H2020 SME Instrument program, under Grant Agreement no. 849628, project OMNIUS, the grant 2016-DI-095 from the Secretaria d'Universitats i Recerca del Departament d’Economia i Coneixement de la Generalitat de Catalunya, the Ramon y Cajal Fellowship RYC-2014-16831 and the CERCA Programme / Generalitat de Catalunya.

\bibliographystyle{unsrt}
\bibliography{sample.bib}

%% main text

\end{document}